\documentclass{aiproc}
\usepackage{graphicx}
\usepackage[utf8]{inputenc}
\usepackage{polski-fonts}
\usepackage{url}
\usepackage{amssymb,amsfonts,amsmath}
\usepackage{tabularx}
\usepackage{algorithmic}
\usepackage[caption=false,font=footnotesize]{subfig}

\makeatletter
\renewenvironment{table}
               {\setlength\abovecaptionskip{10\p@}%
                \setlength\belowcaptionskip{10\p@}%
                \@float{table}}
               {\end@float}
\newcommand{\algorithmicbreak}{\textbf{break}}
\newcommand{\BREAK}{\STATE \algorithmicbreak}
\newcommand{\algorithmiccontinue}{\textbf{continue}}
\newcommand{\CONTINUE}{\STATE \algorithmiccontinue}
\makeatother

\begin{document}

\title{Dynamic Vehicle Routing Problem:\\A Monte Carlo approach}
\titlerunning{DVRP: MC approach}
\author{Michał Okulewicz\inst{1,}\inst{3}, \and Jacek Mańdziuk \inst{1,}\inst{2}}
\authorrunning{Michał Okulewicz et al.}
\institute{Warsaw University of~Technology, Faculty of~Mathematics and Information Science, \\ ul. Koszykowa 75, 00-662 Warsaw, Poland
\and
School of~Computer Engineering, Nanyang Technological University,\\
Block N4, Nanyang Avenue, Singapore 639798
\and
Institute of~Computer Science, Polish Academy of~Sciences,\\ ul. Jana Kazimierza 5, 01-248  Warsaw, Poland
}

\maketitle
\begin{abstract}
\noindent
In this work we solve the~Dynamic Vehicle Routing Problem (DVRP).
DVRP is a~modification of~the~Vehicle Routing Problem,
in which the~clients' requests (cities) number and location might not
be known at~the~beginning of~the~working day
Additionally, all requests must be served during one working day
by a~fleet of~vehicles with limited capacity.  
In this work we propose a~Monte Carlo method (MCTree),
which directly approaches the~dynamic nature
of arriving requests in~the~DVRP.
The method is also hybridized (MCTree+PSO) with our previous Two--Phase Multi-swarm Particle Swarm Optimization (2MPSO)
algorithm.

Our method is based on two assumptions. First, that we know
a bounding rectangle of~the~area in~which the
requests might appear. Second, that the~initial requests'
sizes and frequency of~appearance are representative
for the~yet unknown clients' requests.
In order to~solve the~DVRP we divide the~working day
into several time slices in~which we solve
a static problem. In~our Monte Carlo approach
we randomly generate the~unknown clients' requests
with uniform spatial distribution over the~bounding rectangle and requests'
sizes uniformly sampled from the~already known requests' sizes.
The solution proposal is constructed with
the application of~a~clustering algorithm
and a~route construction algorithm.

The MCTree method is tested on a~well established
set of~benchmarks proposed by Kilby et al.~\cite{DVRP:Study,DVRP:Benchmark} and
is compared with the~results achieved by
applying our previous 2MPSO~\cite{DVRP:2MPSO} algorithm and other literature results.
The proposed MCTree approach achieves
a better time to~quality trade--off then plain heuristic
algorithms. Moreover, a~hybrid MCTree+PSO approach 
achieves better time to~quality trade--off then 2MPSO
for small optimization time limits, making the~hybrid a~good
candidate for handling real world scale goods delivery
problems.

\end{abstract}

\section{Introduction}
The static Vehicle Routing Problem (VRP) has been introduced in~1959 by Dantzig and~Ramser~\cite{VRP:Introduction}. The~VRP is a~generalization of~the~Traveling Salesman Problem, where the~weighted nodes representing requests are served by the~fleet of~vehicles with limited capacity. Since its introduction the~problem has received much attention
in the~literature (e.g. Fisher and Jaikumar 1981~\cite{VRP:FisherJakumar81},
Christofides and Beasley 1984~\cite{VRP:ChristofidesBeasley84}, Taillard 1993~\cite{VRP:Taillard93},
Toth and Vigo 2001~\cite{VRP}). Moreover, the~problem itself
has been generalized in~various ways in~order to~model certain real-world scenarios, such as: VRP with Time Windows~\cite{solomon1987algorithms}, VRP with Pickup and Delivery~\cite{righini2000approximation},
Stochastic VRP~\cite{laporte1998solving}, VRP with Heterogeneous Fleet~\cite{prins2002efficient} etc.
One of~the~most recent generalizations is the~Vehicle Routing Problem with Dynamic Requests,
which is more often simply called the~Dynamic Vehicle Routing Problem (DVRP).

Although the~quality of~the~DVRP solutions has been successfully improved in~the~recent years by the~two meta-heuristic
based algorithms: Two--Phase Multi-swarm Particle Swarm Optimization (2MPSO)~\cite{DVRP:2MPSO} and Multi-environmental cooperative parallel metaheuristics (MEMSO)~\cite{DVRP:MEMSO}, the~computational effort
of obtaining good solutions might be two high for the~real world scenarios, were one considers several thousand requests and a~few hundred vehicles during one working day.

On the~other hand solutions achieved by a~fast heuristic approach might be of~a~poor quality.
Therefore, in~this paper the~authors investigate a~method of~improving heuristic algorithm
results in~dynamic environment by generating artificial data and assess the~time-to-quality
ratio of~their various configurations.

The rest of~the~paper is organized as follows. Section~\ref{sec:dvrp}
gives the~definition of~the~DVRP considered in~this paper.
Section~\ref{sec:solving} presents some general remarks
for solving dynamic problems and describes the~role of~the~\emph{cut-off time} parameter.
Sections~\ref{sec:generation}, \ref{sec:mctree}, \ref{sec:2mpso} and~\ref{sec:mctree+pso}
describe generating of~artificial requests procedure and the~MCTree, 2MPSO and MCTree+PSO algorithms used to~solve
the problem, respectively.
Section~\ref{sec:results} presents the~results obtained by the~proposed method
and compared with literature results. Finally, Section~\ref{sec:discussion}
concludes the~paper.

\begin{table}[!ht]
\caption{Symbols used in~this paper}
\centering
\begin{tabularx}{\textwidth}{Xll}
Symbol & Type & Description \\\hline
\multicolumn{3}{c}{Dynamic Vehicle Routing Problem}\\\hline
$C$ & Series & Clients requests\\
$C'_t$ & Series & Artificial clients requests generated at~time $t$\\
$c_i$ & Client & $i$th client (quadruple of~size, location, duration and time)\\
$V$ & Series & Fleet of~vehicles \\
$v_i$ & Vehicle & $i$th vehicle\\
$n$ & $\mathbb{Z_+}$ & Number of~vehicles\\
$m$ & $\mathbb{Z_+}$ & Number of~clients\\
$m_t$ & $\mathbb{Z_+}$ & Number of~clients known at~the~time $t$\\
$m'_t$ & $\mathbb{Z_+}$ & Number of~artificial clients generated at~the~time $t$\\
$l_i$ & $\mathbb{R}^2$ & Location of~the~$i$th client\\
$l_i.x$ & $\mathbb{R}$ & $x$ coordinate of~the~$i$th client\\
$l_i.y$ & $\mathbb{R}$ & $y$ coordinate of~the~$i$th client\\
$L'_t$ & Series & Artificial clients requests' locations generated at~time $t$\\
$u_i$ & $\mathbb{R_+}$ & Cargo unloading time for the~$i$th client\\
$\mathbf{{{\bar{u}}_{m_t}}}$ & $\mathbb{R_+}$ & Average unloading time for the~first $m_t$ clients\\
$s_i$ & $\mathbb{R_+}$ & Size of~the~$i$th request\\
$S'_t$ & Series & Artificial clients requests' sizes generated at~time $t$\\
$t_i$ & $\mathbb{R}$ & Time when the~request of~the~$i$th client becomes known\\
$l_0$ & $\mathbb{R}^2$ & Location of~the~depot\\
$\rho(i,j)$ & $\mathbb{R}$ & Distance between $l_i$ and $l_j$\\
$t_{start}$ & $\mathbb{R}$ & Depot opening time\\
$t_{end}$ & $\mathbb{R}$ & Depot closing time\\
$cap$ & $\mathbb{R_+}$ & Capacity of~the~vehicle\\
$sp$ & $\mathbb{R_+}$ & Speed of~the~vehicle\\
$\mathbf{r_i}$ & Series & Route of~the~$i$th vehicle (indexes of~subsequent locations) \\
$m_i$ & $\mathbb{Z_+}$ & Number of~locations assigned to~$i$th vehicle\\
$arv_{r_{i,j}}$ & $\mathbb{R}$ & Arrival time of~the~$i$th vehicle at~its $j$th client\\
$t$ & $\mathbb{R}$ & Time ($\forall_t t_{start} \leq t \leq t_{end}$)\\
\hline
\multicolumn{3}{c}{Problem Solving Framework}\\\hline
$T_{CO}$ & $[0,1]$ & New requests arrival \textit{cut-off time}; fraction of~the~working day\\
\hline
\multicolumn{3}{c}{Particle Swarm Optimization}\\\hline
$n$ & $\mathbb{Z}$ & Search space dimension\\
$t$ & $\mathbb{Z}$ & Iteration number \\
$x_t$ & $\mathbb{R}^n$ & Particle's location in~$t$th iteration\\
$v_t$ & $\mathbb{R}^n$ & Particle's velocity in~$t$th iteration\\
$g$ & $\mathbb{R}$ & Best neighbor location attraction factor\\
$l$ & $\mathbb{R}$ & Historically best location attraction factor\\
$a$ & $\mathbb{R}$ & Particles' velocity inertia coefficient\\
$u^{(i)}$ & $\mathbb{R}^n$ & a~random vector with $n$-dimensional uniform distribution\\
$x^i_{best}$~ & $\mathbb{R}^n$ & the~best location visited by the~$i$th particle\\
$x^{n[i]}_{best}$ & $\mathbb{R}^n$ & the~best location visited by the~neighbors of~the~$i$th particle\\
\end{tabularx}
\vspace{-1em}
\end{table}

\section{DVRP Definition}
\label{sec:dvrp}

In the~class of~Dynamic Vehicle Routing Problems discussed in~this
article one considers a~fleet $V$ of~$n$ vehicles and a~series $C$
of $m$ clients (requests) to~be served (a cargo is to~be delivered
to them).

The fleet of~vehicles is homogeneous. Vehicles have identical capacity $cap$ and the~same speed\footnote{In all benchmarks
used in~this paper $sp$ is equal to~one distance unit per one
time unit.} $sp$.

The cargo is taken from a~single depot which has a~certain
location $l_0$ and working hours from $t_{start}$ to~$t_{end}$.

Each client $c_i, i=1,\ldots, m$ has a~given location $l_i$,
time $t_i$ (a point in~time
when the~$i$th request becomes available ($t_{start} \leq t_i \leq t_{end}$)),
time $u_i$ (time required to~unload the
cargo), and size of~the~request $s_i$ ($s_i
\leq cap$).

A travel distance $\rho(i,j)$ is the~Euclidean distance between
$l_i$ and $l_j$ on the~$\mathbb{R}^2$ plane,
$i,j=0,\ldots, m$.

The route $\mathbf{r_i}$ of~vehicle $v_i$ is a~series of~$r_{i,j}$ indexes
of clients,
where the~first $r_{i,1} = 0$ and last $r_{i,m_i} = 0$ are the~identifiers of~the~depot.
Additionally, the~series of~$arv_{r_{i,j}}$ time points defines vehicle's time of~arrival at~those locations.

The goal of~the~DVRP is to~serve the~clients (requests),
according to~their defined time and size constraints, with minimal total cost
(travel distance). Formally, the~optimization goal can be written as:
\begin{equation}
\label{eq:goal}
    \text{min} \sum\limits_{i=1}^n \sum\limits_{j=2}^{m_i} \rho(r_{i,j-1},r_{i,j})
\end{equation}
The feasible solution must fulfill the~following constrains:
\begin{itemize}
\item Each client is served by only one vehicle (requests may not be divided):
\begin{equation}
\label{eq:undivided}
    \forall_{j \in \lbrace 1,2,\ldots m \rbrace} \exists!_{i \in \lbrace 1,2,\ldots n \rbrace} j \in \mathbf{r_i}
\end{equation}
\item the~vehicle cannot arrive at~the~location of~the~next request before the~previous one is served:
\begin{equation}
\label{eq:arriv}
    \forall_{i \in \lbrace 1,2,\ldots n \rbrace}\forall_{j \in \lbrace 2, 3 \ldots m_i \rbrace} arv_{r_{i,j}} \geq arv_{r_{i,j-1}} + u_{r_{i,j-1}} + \rho(r_{i,j-1},r_{i,j}) \\
\end{equation}
\item the~vehicle cannot leave the~previous location before the~next one is known:
\begin{equation}
\label{eq:leave}
    \forall_{i \in \lbrace 1,2,\ldots n \rbrace}\forall_{j \in \lbrace 2, 3 \ldots m_i \rbrace} arv_{r_{i,j}} \geq t_{r_{i,j}} + \rho(r_{i,j-1},r_{i,j}) \\
\end{equation}
\item the~vehicle must return to~the~depot before closing time:
\begin{equation}
\label{eq:return}
    \forall_{i \in \lbrace 1,2,\ldots n \rbrace} arv_{r_{i,m_i}} \leq t_{end} \\
\end{equation}
\item the~vehicle must not leave the~depot before opening time:
\begin{equation}
\label{eq:start}
    \forall_{i \in \lbrace 1,2,\ldots n \rbrace} arv_{r_{i,1}} \geq t_{start} \\
\end{equation}
\item the~capacity of~the~vehicle must not be exceeded between two subsequent visits to~the~depot:
\begin{equation}
\label{eq:capacity}
\forall_{i \in \lbrace 1,2,\ldots n \rbrace} \forall_{j_1 < j_2 < j_3 \in \lbrace 1, 2 \ldots m_i \rbrace} r_{i,j_1} = 0 \wedge r_{i,j_3} = 0 \wedge r_{i,j_2} \neq 0 \Rightarrow \sum\limits_{j=j_1+1}^{j_3-1}s_{r_{i,j}} \leq cap \\
\end{equation}
\end{itemize}

\section{Solving the~DVRP}
\label{sec:solving}
\begin{figure}[!ht]
\centering
	\includegraphics[width=0.9\textwidth]{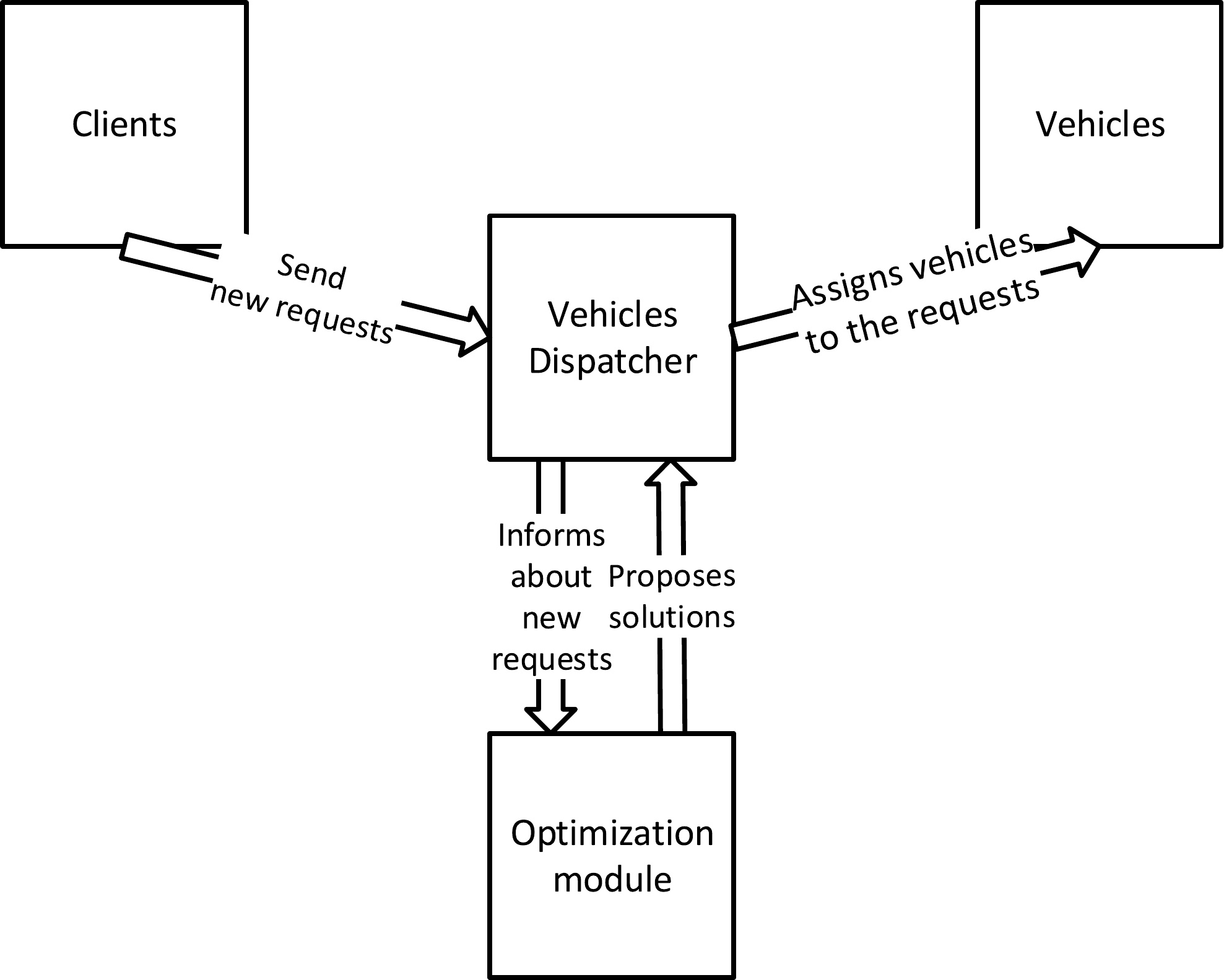}
	\caption{The general framework for solving DVRP}
	\label{fig.framework}
\end{figure}

There are two general approaches to~solving dynamic optimization
problems. In~the~first one the~optimization algorithm is notified every
time there is a~change in~the~problem instance. In~the~second
approach time is divided into a discrete slices and the~algorithm is
run once for each time slice. Furthermore, the~problem instance is
considered "frozen" during the~whole time slice, i.e. any potential
changes introduced during the~current time slot are handled in~the
next algorithm's run (in the~subsequent time slice period).

Another important property of a given dynamic problem is a \emph{degree
of dynamism} (i.e. the ratio of the amount of information unknown
at the beginning of the optimization process to the total amount
of information provided during the optimization). In the case
of the DVRP this property may be measured by
the number of requests unknown at the beginning of the working
day divided by the total number of requests in the given problem instance.
In the DVRP the \emph{degree of dynamism} is a function of a
parameter called \emph{cut--off time} discussed further in this section.

\subsection{General DVRP Framework}

Regardless of~the~approach and the \emph{degree
of dynamism},
the~whole system for solving the~DVRP
may be depicted as in~Fig.~\ref{fig.framework}, where
the Optimization module (topic of~this paper)
acts as a~service for the~Vehicles Dispatcher (VD)
(a person or a~system). Optimization module presents VD with the~best found solution
on the~basis of~{the provided} data.

\subsection{Cut--off Time}
As mentioned before, an important DVRP parameter which has a~direct impact on the
\emph{degree of~dynamism} of~a~given problem instance, is the~\emph{cut--off time} factor.
It defines the~part of~the~requests set that is known at~the
beginning of~the~working day. In~the~real (practical) situations the
requests received after this time threshold are moved to~be served
at the~subsequent working day.
In the~one-day-horizon simulations presented in~this paper (as well as in
practically all other papers referring to~Kilby et al.'s
benchmarks~\cite{DVRP:Benchmark}) the~requests located after the~\emph{cut--off time} limit are simply treated as being known at~the~beginning of
the current day - they compose an~initial instance of~the~DVRP being solved.

\subsection{Benchmark Configuration and Optimization Settings}
In this article we follow the~approach with
dividing working day into discrete time slices
 and running optimization algorithm
in each time slice for a~"frozen" static VRP instances.
This approach in~the~context of
the DVRP, was proposed by Kilby et al.~\cite{DVRP:Study}
{and has been followed in~our previous work with
the Particle Swarm Optimization algorithm~\cite{DVRP:2PSO,DVRP:2MPSO}.}

In order to
assure a~possibility of~direct comparison of~the~obtained results with our previous
work and with other literature results~\cite{DVRP:DAPSO,DVRP:MAPSO,DVRP:MEMSO,DVRP:GA:TS,DVRP:Ants,DVRP:GA2014,DVRP:ACOLNS},
the  \emph{cut--off time} is set in~the
middle of~a~depot's working hours.

\section{Requests generation}
\label{sec:generation}
In order to~approach the~dynamic nature of~the~problem,
in each time slice occurring prior to~the~\emph{cut-off time} a~set of~randomly generated requests 
is added to~the~currently known ones.
Thus, the~(D)VRP is solved from scratch in~the~subsequent time slices
with decreasing amount of~unknown requests and increasing amount
of fixed assignments of~vehicles to~the~requests (due to~advancing time).
In each time slice it is assumed that
we know the~maximum spatial range {in} which the~requests
may be located {(see Eq.~(\ref{eq:location.artificial.requests}))} and that the~frequency of~the~requests arrival
during the~remaining part of~the~working day will be the~same as up till
{the} current point in~time (see Eq.~(\ref{eq:no.artificial.requests})).
Therefore, new requests are generated uniformly in~time
and space and their size is uniformly sampled from the~already known requests {(see Eq.~(\ref{eq:size.artificial.requests}))}.

{$C'_t$ is a~quadruple of~sizes, locations, availability times (all set to~the~current time) and unload times
(all set to~the~average known unload time).
\begin{equation}
C'_t := (
	S'_t,
	L'_t,
	\lbrace{t,\ldots,t}\rbrace,
	\lbrace\mathbf{{{\bar{u}}_{m_t}}},\ldots\mathbf{{{\bar{u}}_{m_t}}}\rbrace)
	\label{eq:artificial.requests}
\end{equation}
The number {$m'_t$} of~the~generated $C'_t$ requests at~time $t$, is computed as follows:
\begin{equation}
m'_t = m_t\dfrac{T_{CO}(t_{end}-t_{start})+t_{start}-t}{T_{CO}(t_{end}-t_{start})-t_{start}+t}
	\label{eq:no.artificial.requests}
\end{equation}}
The size of~the~requests is sampled uniformly from the~known requests' sizes:
\begin{equation}
S'_t \sim U(\lbrace s_1, s_2, \ldots, s_{m_t} \rbrace)
	\label{eq:size.artificial.requests}
\end{equation}
The location of~the~requests is generated uniformly over the~bounding rectangle
of all of~the~requests:
\begin{equation}
L'_t \sim U(
	[
	\min\limits_{i_1 \in \lbrace 1, 2, \ldots, m \rbrace}(l_{i_1}.x),
	\max\limits_{i_2 \in \lbrace 1, 2, \ldots, m \rbrace}(l_{i_2}.x)
	]
	\times
	[
	\min\limits_{i_3 \in \lbrace 1, 2, \ldots, m \rbrace}(l_{i_3}.y),
	\max\limits_{i_4 \in \lbrace 1, 2, \ldots, m \rbrace}(l_{i_4}.y)
	])
	\label{eq:location.artificial.requests}
\end{equation}
{Uniform distribution models the~assumption of~knowing
only the~spatial boundaries in~which new requests can appear.}

\begin{figure}[ht]
\begin{algorithmic}[1]
    \STATE{}
    \COMMENT{$E$ a~set of~weighted edges, weights represent the~distances between nodes}
    \STATE{}
    \COMMENT{$V$ a~set of~weighted nodes, weights represent the~size of~the~requests}
    \STATE{}
    \COMMENT{$CAPACITY$ a~constant representing the~capacity of~a~single vehicle}
    \STATE{$T_{clusters} \Leftarrow CreateSeparateTrees(V)$} \label{alg.line:initial.clusters}
    \STATE{$E_{sort} \Leftarrow SortByWeightInAscendingOrder(E)$} \label{alg.line:edges}
    \FORALL{$(v_1, v_2) \in E_{sort}$}
        \IF {$Tree(v_1) \neq Tree(v_2)$ \label{alg.line:acyclic.rule}} 
	        \IF {$Weight(Tree(v_1)) + Weight(Tree(v_2)) \leq CAPACITY$ \label{alg.line:capacity.rule}}
		        \IF {$Distance(v_1,v_2) \leq Distance(v_1,Depot)$ \AND $Distance(v_1,v_2) \leq Distance(v_2,Depot)$ \label{alg.line:small.distance.rule}}
        		    \STATE{$T_{clusters} \Leftarrow T_{clusters} \setminus \lbrace Tree(v_1), Tree(v_2) \rbrace$}
		            \STATE{$T_{clusters} \Leftarrow T_{clusters} \cup \lbrace Tree(v_1) \cup Tree(v_2) \rbrace$}
    		    \ENDIF
	        \ENDIF
        \ENDIF
    \ENDFOR
\end{algorithmic}
\caption{Pseudocode of~the~modified Kruskal algorithm for solving a~capacitated clustering task.}
\label{alg:spanning}
\end{figure}

\begin{figure}[ht]
\begin{algorithmic}[1]
    \WHILE{\TRUE}
	    \FORALL{$j_1 \in \lbrace 2,3,\ldots,m_i-1\rbrace$}
	   		\IF {$c_{r_{i,j_1}}$ is fixed \label{alg.line:non.fixed}}
	   			\CONTINUE
	   		\ENDIF
	    	\FORALL{$j_2 \in \lbrace j_1+1,j_1+2,\ldots,m_i\rbrace$}
	    		\IF {$\rho(r_{i,j_1-1},r_{i,j_1}) + \rho(r_{i,j_2-1},r_{i,j_2})
	    			<
	    			\rho(r_{i,j_1-1},r_{i,j_2-1}) + \rho(r_{i,j_1},r_{i,j_2})
	    			$ \label{alg.line:exchange.rule}} 
	    			\FORALL{$j_3 \in \lbrace j_1,j_1+1,\ldots,j_2-1\rbrace$, $j_4 \in \lbrace j_2-1,j_2-2,\ldots,j_1\rbrace$}
	    				\STATE {$r_{i,j_3} \leftrightarrow r_{i,j_4}$\label{alg.line:reverse}}
	    				\IF {$j_4 \leq j_3$}
	    					\BREAK
	    				\ENDIF
	    			\ENDFOR
	    		\ENDIF
	    	\ENDFOR
	    \ENDFOR
   		\IF {nothing to~optimize}
	   	\BREAK
	   	\ENDIF
    \ENDWHILE
\end{algorithmic}
\caption{Pseudocode of~the~2-OPT algorithm for optimizing the~route $\mathbf{r_i}$
of the~$i$th vehicle.}
\label{alg:routing}
\end{figure}

\begin{figure}[ht]
	\centering
	\includegraphics[width=0.8\textwidth]{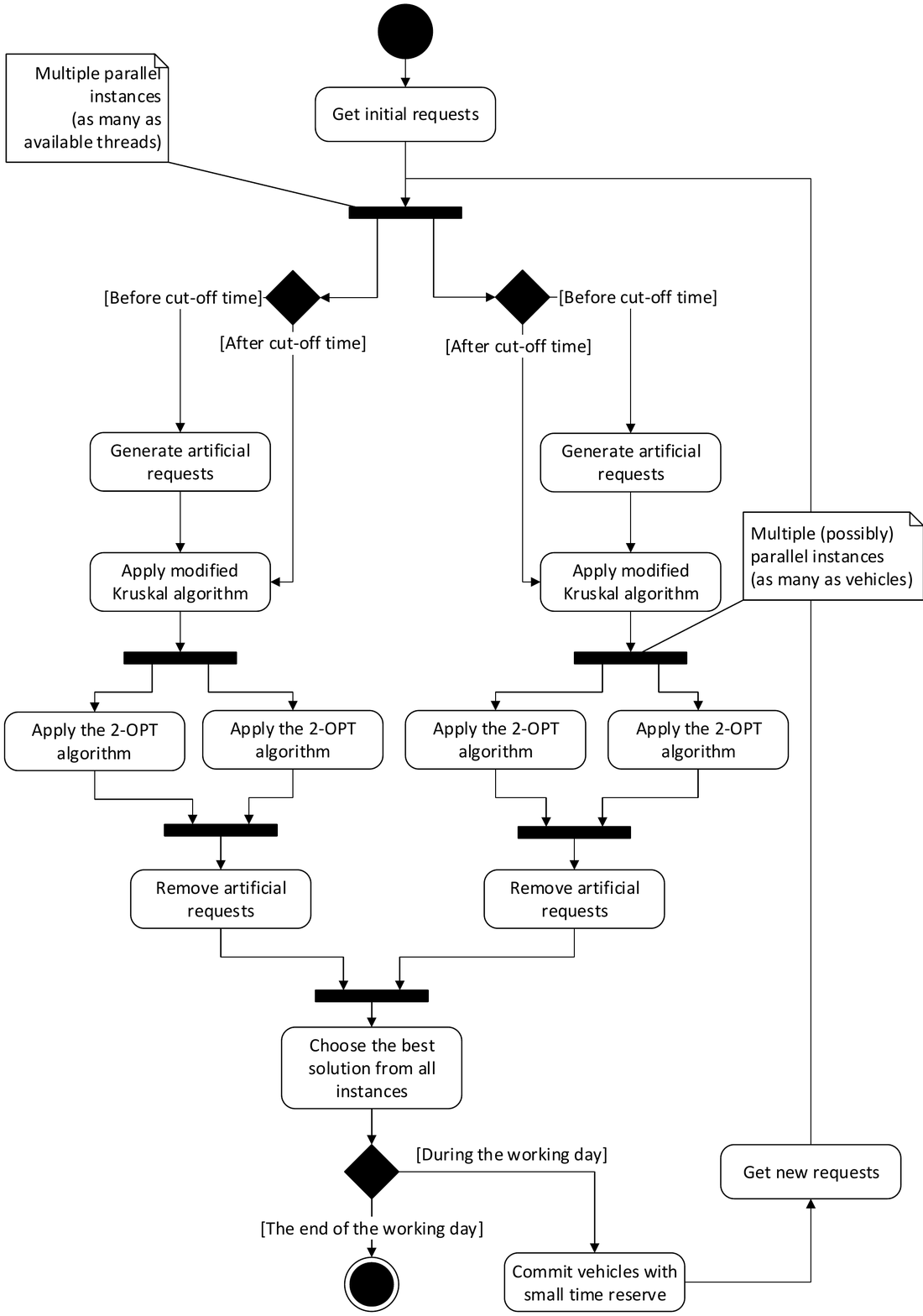}
	\caption{Activity diagram of~the~MCTree algorithm.}
	\label{fig:mctree.activity}
\end{figure}

\section{MCTree algorithm}
\label{sec:mctree}

The~MCTree algorithm for solving the~(D)VRP is a~sequence of~two heuristic algorithms.
In the~first step, a capacitated clustering problem is solved
by the usage of a~modified version of~the~Kruskal algorithm~\cite{TSP:Kruskal},
thus creating a requests-to-vehicles assignment.
In the~second step, the partially random routes of the~vehicles
are optimized with the~2--OPT algorithm~\cite{TSP:2OPT}.
In that step routes are optimized independently within each single cluster of the requests.

\subsection{Requests Assignment Optimization (Clustering)}
The goal of~the~modified Kruskal algorithm is to~create
a forest with each tree representing the~clients assigned
to one vehicle. Please note, that because of~the~limited sum of~nodes' weights
in each tree the~final forest graph might not be
a sub-graph of~the~minimum spanning tree created by a~standard Kruskal algorithm.

The pseudo-code for the~modified Kruskal algorithm is presented in~Fig.~\ref{alg:spanning}.
There are two modifications made in~the~standard algorithm in~order to~achieve
solutions for the~(D)VRP.
The first modification is in~{the} line~\ref{alg.line:capacity.rule},
stating that two trees cannot be joined if {the} sum of~their nodes' weights would exceed the~capacity
of the~vehicle.
The second modification (line~\ref{alg.line:small.distance.rule}) is an~experimentally
chosen heuristic rule, preventing {the} creation of~{the} routes covering large areas
and (possibly) overlapping with others.

Please note, that creating the~initial clusters (line~\ref{alg.line:initial.clusters}),
takes into account clients with fixed assignment to~a~given vehicle and {a} set of~edges
(sorted in~line~\ref{alg.line:edges}) consists only of~{the} edges between all non-fixed clients
and {the} fixed and non-fixed clients, without the~edges between the~fixed clients
(so the~problem size decreases size after a~certain point of~the~day).

\subsection{Routes Optimization}
The goal of~the~2--OPT algorithm is creation of~an~acceptable quality routes
for each of~the~vehicles. The~2--OPT algorithm checks all pairs of~the~non-fixed
edges for the~possibility of~minimizing the~route length by swapping their ends.

Pseudocode of~the~2-OPT algorithm is presented in~Fig.~\ref{alg:routing}.
The 2-OPT executes as long as the~route is improved by exchanging
two edges (line \ref{alg.line:exchange.rule}) and the~part of~the~route
between the~exchanged edges is reversed(line \ref{alg.line:reverse}).

The algorithm had to~be slightly changed in~order to~incorporate
the dynamic nature of~the~optimized problem {(i.e. the~initial part
of the~route becomes fixed)}.
{Instead of~iterating over the~whole route,} the~optimization starts from the~first non-fixed
client $c_{r_{i,j_1}}$ of~the~solution for a~given vehicle $\mathbf{r_i}$ (line \ref{alg.line:non.fixed}).

\subsection{Algorithm Architecture}
The generation of~the~artificial requests, the~clustering algorithm and the~route optimization algorithm
are the~main building blocks of~the~MCTree method.

The activity diagram of~the~MCTree algorithm is depicted in~Fig.~\ref{fig:mctree.activity}.
In each time slice the~algorithm is run in~separate multiple parallel instances (by default
as many as available real or virtual CPUs: 8 in~the~case of~Intel Core i7 used in~this study).
In each of~the~parallel processes the~artificial requests are generated independently.
After {the} generation of~the~requests, first the~Kruskal algorithm and then the~2-OPT algorithm
are applied. Finally, the~artificial requests are removed from the~solution and the~best found {set of} routes
among the~separate parallel instances {is} chosen. Subsequently the~algorithm advances to~the~next time slice.

Please note, that the~generation of~requests is an~optional step
and can be omitted ({such approach is denoted as the~Tree algorithm}).
Also, both the~Kruskal and 2-OPT algorithm
can be replaced {(or enhanced)} with another clustering and route optimization
algorithms (as will be the~case in~the~approach using Particle
Swarm Optimization algorithm (2MPSO)){, which is discussed in~the~next section}.

\section{2MPSO algorithm}
\label{sec:2mpso}
The Two--Phase Multi-swarm Particle Swarm Optimization (2MPSO)
algorithm for the DVRP has been introduced by the~authors in~2014~\cite{DVRP:2MPSO}.
In the~2MPSO algorithm a~continuous optimization Particle Swarm Optimization (PSO)
algorithm is used for the requests clustering and the vehicles' routes optimization.
Therefore, the~authors have proposed a~continuous
encoding for both the~requests-to-vehicle assignment and requests ordering task.

\subsection{Continuous DVRP Encoding}

The assignment is encoded as a~flattened (one dimensional) array of~requests' cluster centers.
Each vehicle is associated with a~set of~such cluster centers, thus all requests belonging
to the~given clusters are assigned to~that vehicle. In~order to~evaluate the~assignment
as a~set of~vehicles' routes, each set of~clients assigned to~one vehicle is treated
as a~random route and optimized with 2--OPT before computing the~value of~the~objective function.

The route ordering for a~given vehicle is achieved by sorting the~indexes
of the~vector of~the~length equal to~the~number of~requests assigned to~that vehicle.
The rank of~the~index corresponding to~the~given requests is defined
by the~rank of~its value in~the~vector.

Creation of~that continuous encodings (together with assignment-to-route conversion)
allowed for a~direct application of~the~PSO algorithm to~the~DVRP.

\subsection{Particle Swarm Optimization}
\label{sec:pso} PSO is an~iterative global optimization
meta-heuristic method proposed in~1995 by Kennedy and
Eberhart~\cite{PSO:Introduction} and further studied and developed
by many other researchers,
e.g.,~\cite{PSO:Inertia,PSO:Modified,PSO:Params}. In
short, PSO utilizes the~idea of~swarm intelligence to~solve hard
optimization tasks. The~underlying idea of~the~PSO algorithm
consists in~maintaining the~swarm of~particles moving in~the~search
space. For each particle the~set of~neighboring particles which communicate
their positions and function values to
this particle is defined. Furthermore, each particle maintains its current
position and velocity, as well as remembers its historically best
(in terms of~solution quality) visited location. More precisely, in~each
iteration $t$, each particle $i$ updates its position $x^i_t$ and
velocity $v^i_{t}$ according to~the~following formulas for the position
and velocity update.

The~position of~a~particle is updated according to the~following equation:
\begin{equation}
    x^i_{t+1} = x^i_t + v^i_t.
\end{equation}
In~our implementation of~the~PSO (based on~\cite{Algorithm:PSO2011}
and~\cite{PSO:Inertia}) the~velocity $v^i_{t}$ of~a~particle $i$ is
updated according to~the~following rule:
\begin{equation}
    v^i_{t+1} =  u^{(1)}_{U[0;g]} (x^{n[i]}_{best} - x^i_t) +  u^{(2)}_{U[0;l]} (x^i_{best} - x^i_t) + a\cdot v^i_t \\ \label{eq:PSO_vi}
\end{equation}
where
\begin{itemize}
    \item $g$ is a~neighborhood attraction factor,
    \item $x^{n[i]}_{best}$ represents the~best position (in terms of~optimization) found hitherto by the~particles belonging to~the~neighborhood of~the~$i$th particle,
    \item $l$ is a~local attraction factor,
    \item $x^i_{best}$ represents the~best position (in terms of~optimization) found hitherto by particle $i$,
    \item $a$ is an~inertia coefficient,
    \item $u^{(1)}_{U[0;g]}$, $u^{(2)}_{U[0;l]}$ are random vectors with uniform
    distribution from the~intervals $[0,g]$ and $[0,l]$,
    respectively.

\end{itemize}

\subsection{Similarities to~MCTree}
The general structure of~the~2MPSO algorithm is similar to~that of~the~MCTree:
\begin{itemize}
	\item in~each time step a~static instance is optimized in~independent parallel instances of~the~algorithm,
	\item First, the~requests assignment problem is solved (in the~case of~2MPSO by solving continuous clustering problem with the~PSO algorithm),
	\item Subsequently, the~routes are optimized for each of~the~vehicles independently (also by applying PSO to~continuous route order encoding),
	\item Finally, the~best solution is chosen among the~parallel instances of~the~algorithm.
\end{itemize}

\subsection{Differences with MCTree}
The main difference between the~2MPSO and MCTree (apart from the~continuous vs. discrete optimization) is the~method of~approaching the~dynamic nature
of the~DVRP problem. In~the~MCTree the~problem is solved from scratch
in each time slice, with a~sort of~safety buffer for vehicles' capacities
created by generation of~artificial requests.
In the~2MPSO the~solution from the~previous time slice is used to~generate
a few of~the~solutions in~the~initial population of~PSO and the~search space
is centered around such a~solution. Therefore, 2MPSO relies on the~knowledge
transfer between the~time slices.

\subsection{Heuristic Algorithms}
In the~2MPSO both Kruskal based clustering and 2--OPT algorithm are heavily used.
Kruskal algorithm is used for finding one of~the~initial solutions in~the~PSO swarm.
2--OPT algorithm is used for creating the~routes from the~proposed assignments,
allowing for evaluating requests-to-vehicles assignment as the~solutions of~the~VRP.

Please refer to~the~\cite{DVRP:2MPSO} and~\cite{DVRP:2PSO} for more detailed
description and parameter setting of~the~2MPSO. Also, please note that
in the~experiments in~this paper the~PSO will be used only to~improve
the requests-to-vehicles assignment (such approach has a~better computations
time to~quality pay off).

\section{MCTree+PSO algorithm}
\label{sec:mctree+pso}
The MCTree+PSO algorithm switches from optimizing with the~MCTree to~2MPSO,
when there are no more unknown requests (i.e. after the~\emph{cut--off time}).
The idea of~switching after the~\emph{cut--off time}
comes from the~fact, that that time MCTree changes into a~plain heuristic
Tree algorithm as there are no more artificial requests to~generate.
Additionally, the~problem to~optimize at~that time of~the~working
day usually becomes smaller, as at~least part of~the~requests-to-vehicles assignment
is already fixed, because the~vehicles had to~start to~deliver the~cargo.
Therefore, instead of~a~plain heuristic Tree algorithm the~2MPSO is used as an~optimizer
in the~second part of~a~working day.

\section{Results}
\label{sec:results}
\begin{table*}[ht]
\centering \caption{Comparison of~the~minimum and the~average total routes length achieved by the~Tree, MCTree and MCTree with 2MPSO after \emph{cut-off time} and 2MPSO
with the~average computation time of~about 20 seconds per benchmark.
The numbers denote the~number of~time slices (TS), parallel instances (P) and fitness function evaluations (FFE).
In the~case of~the~PSO, the~FFE is presented as the~number of~particles multiplied by the~number of~iterations.} \label{tab:comparable}
\begin{center}
\resizebox{\textwidth}{!}{
\begin{tabular}{|r||r|r|r|r|r|r|r|r|}
\hline
 &
    \multicolumn{2}{c|}{Tree} &
    \multicolumn{2}{c|}{MCTree} &
    \multicolumn{2}{c|}{MCTree+PSO} &
    \multicolumn{2}{c|}{2MPSO} \\
&
    \multicolumn{2}{c|}{$200TS \times 8P \times 1FFE$ } &
    \multicolumn{2}{c|}{$200TS \times 8P \times 1FFE$} &
    \multicolumn{2}{c|}{$20TS \times 8P \times 1FFE$+} &
    \multicolumn{2}{c|}{$40TS \times 8P \times (4\times28)FFE$} \\
    &
    \multicolumn{2}{c|}{} &
    \multicolumn{2}{c|}{} &
    \multicolumn{2}{c|}{$20TS \times 8P \times (7\times49)FFE$} &
    \multicolumn{2}{c|}{} 
 \\\hline
  & Min & Avg
  & Min & Avg
  & Min & Avg
  & Min & Avg

\\\hline\hline

c50 & 673.34 & 721.51 & 654.69 & 700.49 & 621.27 & 677.03 & \textbf{566.98} & \textbf{610.47} \\ \hline 
c75 & 1049.07 & 1117.71 & 1038.80 & 1123.43 & 998.72 & 1066.00 & \textbf{927.22} & \textbf{988.06} \\ \hline 
c100 & 1095.82 & 1193.98 & 1004.15 & 1119.42 & 979.95 & 1066.44 & \textbf{930.33} & \textbf{1038.53} \\ \hline 
c100b & 828.94 & 843.07 & 828.94 & 836.75 & \textbf{823.23} & \textbf{831.72} & 828.63 & 858.75 \\ \hline 
c120 & 1072.86 & 1106.33 & 1078.23 & 1109.12 & \textbf{1068.46} & \textbf{1100.28} & 1071.20 & 1112.88 \\ \hline 
c150 & 1318.78 & 1463.85 & 1269.23 & 1399.93 & 1223.15 & 1323.35 & \textbf{1205.80} & \textbf{1306.79} \\ \hline 
c199 & 1644.67 & 1824.89 & 1571.05 & 1702.24 & 1533.68 & 1601.22 & \textbf{1471.16} & \textbf{1597.98} \\ \hline 
\hline 
f71 & 290.37 & 348.80 & 303.49 & 333.99 & 288.72 & 323.85 & \textbf{278.56} & \textbf{310.31} \\ \hline 
f134 & 12730.29 & 13501.66 & 12719.73 & 13474.06 & \textbf{12134.30} & \textbf{12473.32} & 12377.63 & 12746.26 \\ \hline 
\hline 
tai75a & 1864.10 & 2016.29 & 1929.44 & 2016.40 & 1899.17 & 1980.37 & \textbf{1832.11} & \textbf{1957.01} \\ \hline 
tai75b & 1578.05 & 1631.85 & 1523.07 & 1655.02 & 1515.71 & \textbf{1582.37} & \textbf{1499.58} & 1611.63 \\ \hline 
tai75c & 1614.94 & 1800.66 & 1570.30 & 1704.14 & \textbf{1526.75} & 1657.88 & 1555.36 & \textbf{1642.58} \\ \hline 
tai75d & 1431.89 & 1612.26 & 1430.75 & 1467.25 & \textbf{1426.39} & \textbf{1456.08} & 1444.70 & 1520.84 \\ \hline 
tai100a & 2359.58 & 2651.78 & 2441.47 & 2640.06 & 2344.78 & 2532.85 & \textbf{2311.19} & \textbf{2467.77} \\ \hline 
tai100b & 2324.19 & 2475.94 & 2297.78 & 2446.81 & 2221.67 & 2344.13 & \textbf{2204.54} & \textbf{2323.27} \\ \hline 
tai100c & 1621.59 & 1752.83 & 1619.81 & 1742.85 & 1580.00 & 1676.72 & \textbf{1566.75} & \textbf{1675.89} \\ \hline 
tai100d & 2019.34 & 2195.64 & 1909.52 & 2050.51 & 1888.07 & 2000.07 & \textbf{1789.90} & \textbf{1960.36} \\ \hline 
tai150a & 3599.62 & 3839.13 & \textbf{3555.79} & \textbf{3711.96} & 3607.78 & 3763.44 & 3664.12 & 3904.32 \\ \hline 
tai150b & \textbf{3052.73} & 3377.34 & 3145.64 & 3268.63 & 3070.44 & \textbf{3226.49} & 3104.98 & 3238.59 \\ \hline 
tai150c & 2718.31 & 2867.25 & 2707.18 & 2844.86 & \textbf{2614.59} & \textbf{2725.45} & 2734.77 & 2874.79 \\ \hline 
tai150d & 3194.07 & 3419.10 & 3100.97 & 3352.11 & \textbf{3081.26} & \textbf{3200.29} & 3134.93 & 3247.68 \\ \hline 
tai385 & \textbf{29088.27} & \textbf{31144.04} & 31331.32 & 33037.53 & 31876.05 & 33786.47 & 30122.29 & 32433.18 \\ \hline 
\hline 
sum & 77170.82 & 82905.91 & 79031.35 & 83737.56 & 78324.14 & 82395.82 & \textbf{76622.73} & \textbf{81427.94} \\ \hline 

 \end{tabular}
} \end{center}
\end{table*}

In order to~assess the~performance of~the~algorithms, we use a~well established set of~benchmarks
created by Kilby et al.~\cite{DVRP:Study} by converting static instances of~VRP
used by Christofides~\cite{VRP:ChristofidesBeasley84}, Fisher~\cite{VRP:FisherJakumar81} and Taillard~\cite{VRP:Taillard93}.

While our previously introduced~\cite{DVRP:2MPSO} Two--Phase Multi-swarm Particle Swarm Optimization (2MPSO)
outperformed other algorithms using limit on number of~fitness function evaluations as a~criterion,
in this article we shall focus on comparison of~the~MCTree with the~algorithms which use time as the~limit for the~optimization process:
Genetic Algorithm (GA2007) (by Hanshar et al.~\cite{DVRP:GA:TS}), Ant Colony Optimization with Large Neighborhood Search (ACOLNS) (by Elhassania et al.~\cite{DVRP:ACOLNS}) and Genetic Algorithm (GA2014)
(by Elhassania et al.~\cite{DVRP:GA2014}).

The tests consisted of~running each version of~the~optimization method 30 times
for each of~the~22 benchmark instances with four different approaches:
\begin{itemize}
	\item \textbf{Tree} - in~each parallel instance one solution was computed with clustering and route optimization heuristic algorithms without previous generation of~the~artificial requests,
	\item \textbf{MCTree} - in~each parallel instance  one solution was computed with clustering and route optimization heuristic algorithms on the~set consisting of~real and artificial requests (till the~\emph{cut-off time}),
	\item \textbf{MCTree+PSO} - till the~\emph{cut-off time} in~each parallel instance one solution was computed on the~partially artificial set of~requests, after the~ \emph{cut-off time} the~2MPSO algorithm was used,
	\item \textbf{2MPSO} - 2MPSO algorithm was used for the~whole experiment, no artificial requests were generated.
\end{itemize}

\begin{figure}[ht!]
	\centering
	\includegraphics[width=0.95\textwidth]{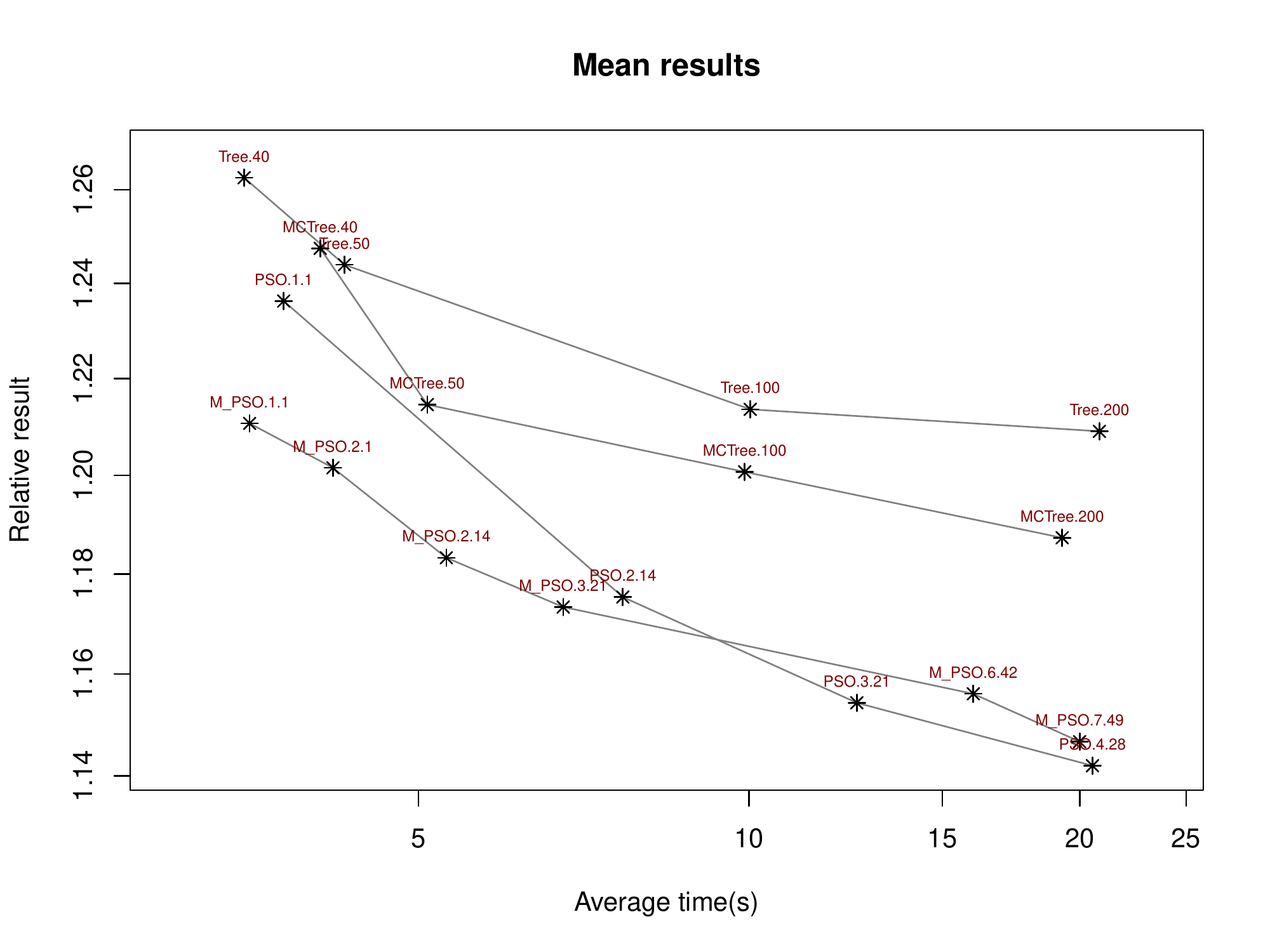}
		\vspace*{-1em}
	\caption{Presentation of~the~result vs. computation time trade-off
	for a~different sets of~parameters for the~MCTree and 2MPSO algorithm.}
	\label{fig.time.2mpso.vs.mctree}
\end{figure}

\begin{figure}[ht]
    \subfloat[Comparison of~the~results achieved by a~Tree algorithm,
	with different number of~time slices.\label{fig:tree}]{%
      \includegraphics[width=0.47\textwidth]{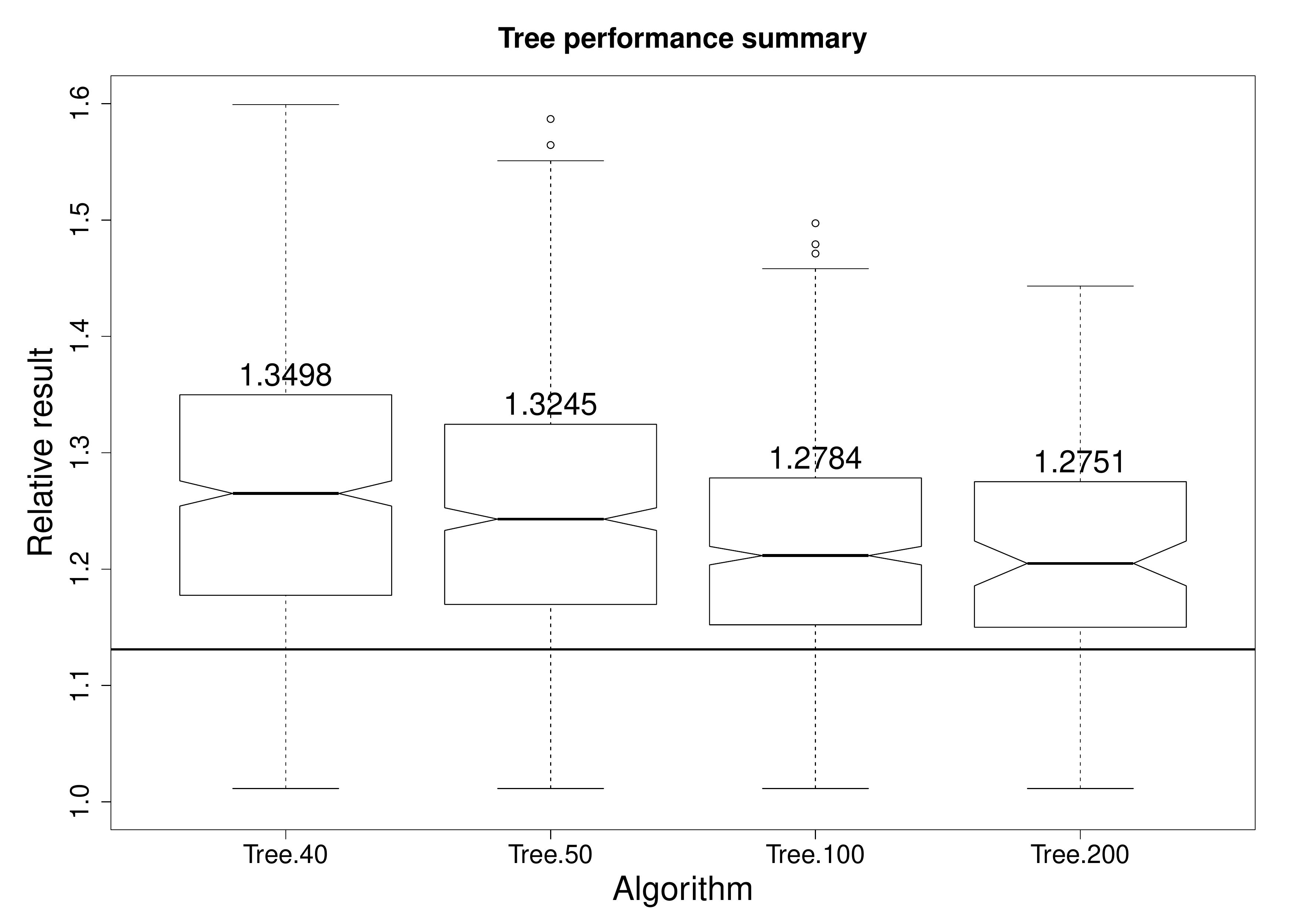}
    }
    \hfill
    \subfloat[Comparison of~the~results achieved by an~MCTree algorithm,
	with different number of~time slices.\label{fig:mctree}]{%
      \includegraphics[width=0.47\textwidth]{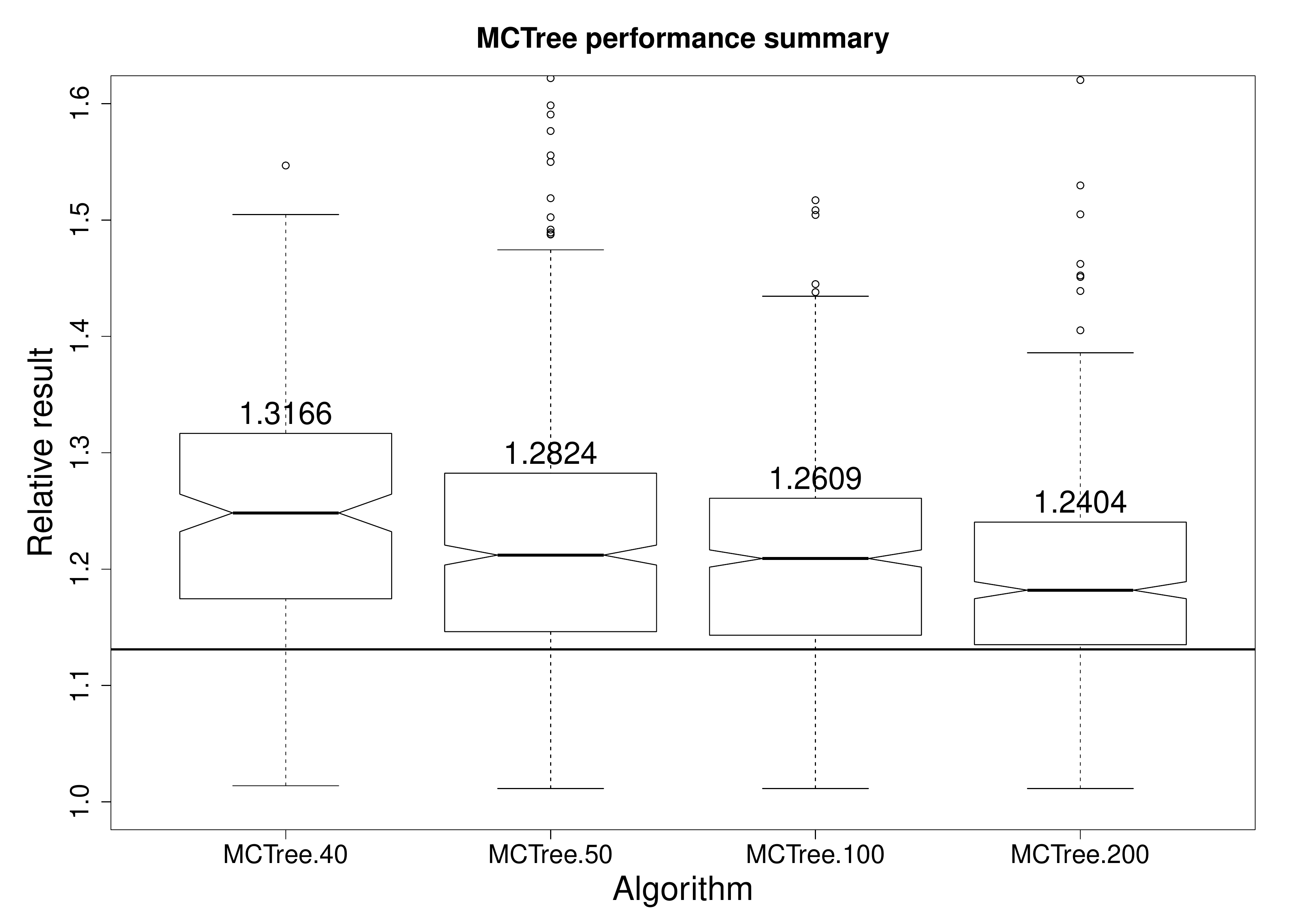}
    }
    \\
    \subfloat[Comparison of~the~results achieved by a~2MPSO algorithm,
	with different swarm size and number of~iterations.\label{fig:mpso}]{%
      \includegraphics[width=0.47\textwidth]{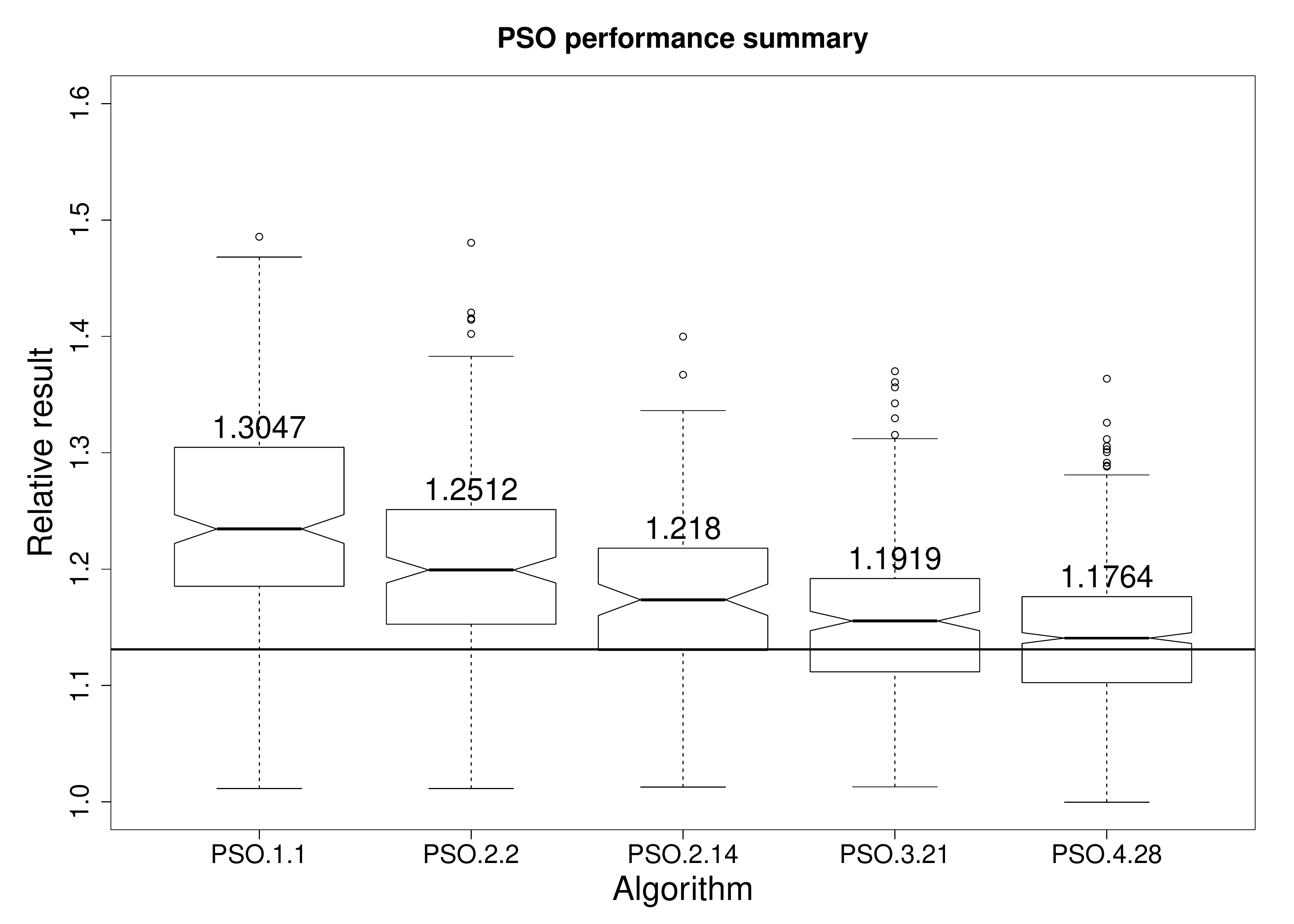}
    }
    \hfill
    \subfloat[Comparison of~the~results achieved by an~MCTree with PSO at~the~end algorithms,
	with different swarm size and number of~iterations.\label{fig:mctreepso}]{%
      \includegraphics[width=0.47\textwidth]{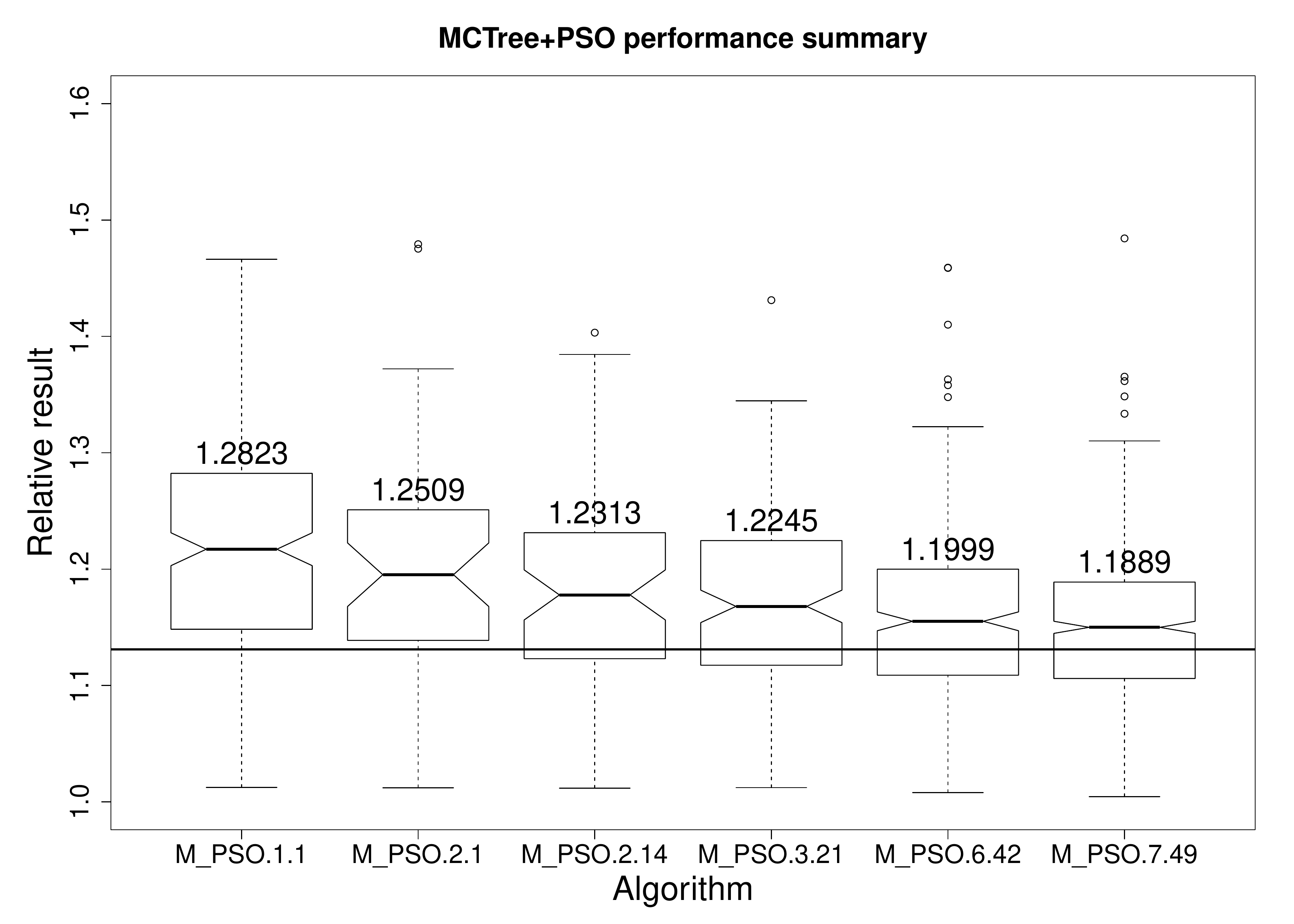}
    }
    \caption{Comparison of~the~performance of~different algorithms.
    The~comparison is done on the~basis of~relative results (i.e.
    results divided by the~best known result for the~given benchmark).
    The~horizontal line represents the~mean value achieved by the
    GA~\cite{DVRP:GA:TS} algorithm.
    Numbers above the~boxplots denote the~3rd quartile of~the~relative results.
    In~each of~the~subplots leftmost and rightmost results were achived by
    the~algorithm computing on average around 4 seconds per experiment and around 20 seconds
    per experiment respectively.}
    \label{fig:algorithm.performance}
\end{figure}

\begin{figure}[ht]
	\includegraphics[width=\textwidth]{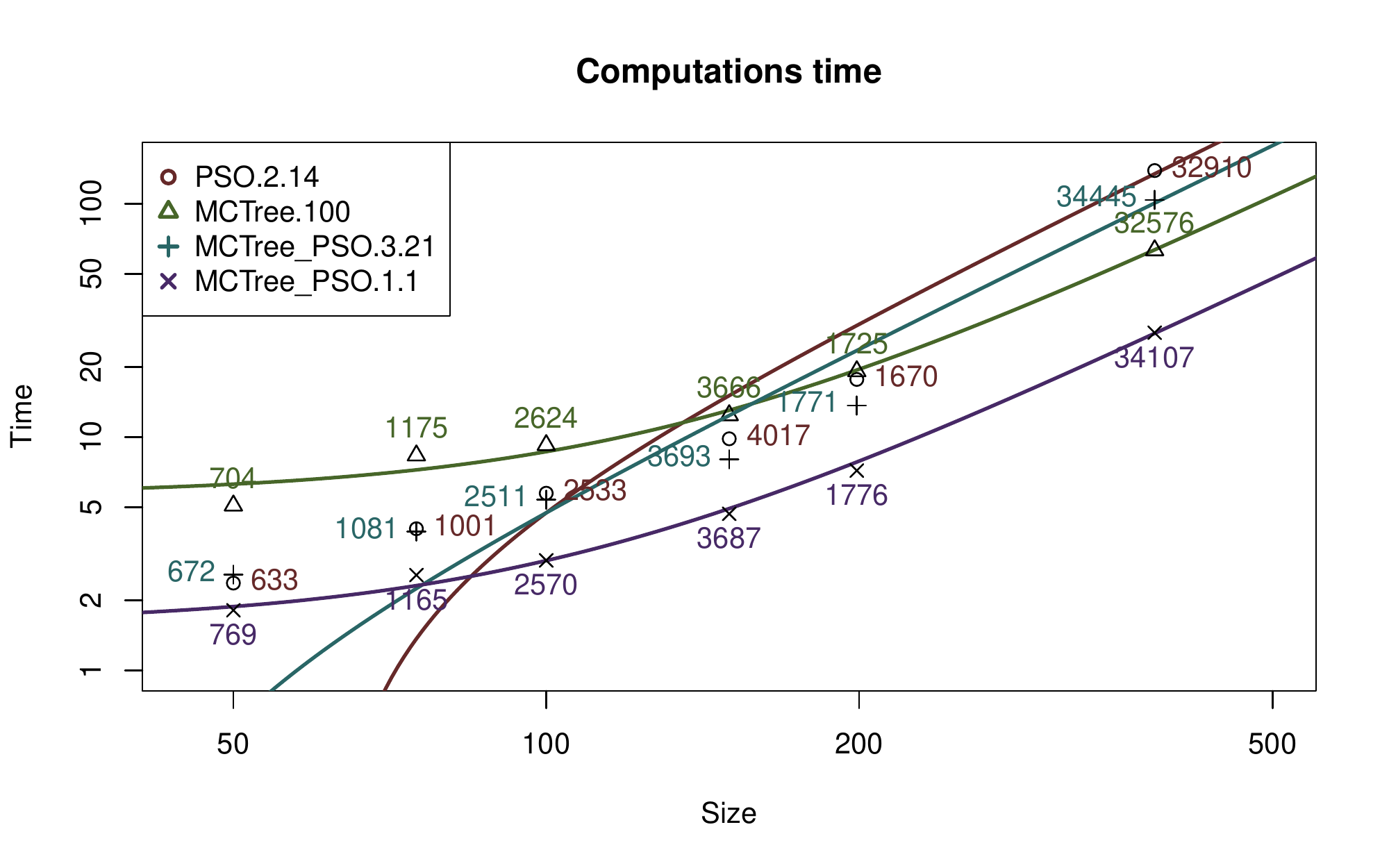}
	\caption{Comparison of~the~computational effort needed for benchmarks
	of different number of~requests. Labels denote the~mean value obtained
	by the~given algorithm for a~chosen benchmark size.
	The regression curves were fitted for the~time as a~linear function of~$m^2log(m)$.}
	
	\label{fig:time.compare}
\end{figure}

\begin{table*}[ht]
\centering \caption{Comparison of~the~minimum and the~average total routes length achieved by the~MCTree with PSO after \emph{cut-off time} approach and the~GA2007, GA2014 and ACOLNS algorithms.
The time limit and the~processor on which the~results
were computed is given for each of~the~algorithms.}
 \label{tab:external}
\begin{center}
\resizebox{\textwidth}{!}{
\begin{tabular}{|r||r|r|r|r|r|r|r|r|}
\hline
 &
    \multicolumn{2}{c|}{$GA2014$~\cite{DVRP:GA2014}} &
    \multicolumn{2}{c|}{$ACOLNS$~\cite{DVRP:ACOLNS}} &
    \multicolumn{2}{c|}{$GA2007$~\cite{DVRP:GA:TS}} &
    \multicolumn{2}{c|}{$MCTree+PSO$} \\
&
    \multicolumn{2}{c|}{1500 seconds} &
    \multicolumn{2}{c|}{1500 seconds} &
    \multicolumn{2}{c|}{750 seconds} &
    \multicolumn{2}{c|}{75 seconds (500 for tai385)} \\
    &
    \multicolumn{2}{c|}{Intel Core i5@2.4GHz} &
    \multicolumn{2}{c|}{Intel Core i5@2.4GHz} &
    \multicolumn{2}{c|}{Intel PentiumIV@2.8GHz} &
    \multicolumn{2}{c|}{Intel Core i7(2nd)@3.4GHz} 
 \\\hline
  & Min & Avg
  & Min & Avg
  & Min & Avg
  & Min & Avg

\\\hline\hline

c50 & 602.75 & 618.86 & 601.78 & 623.09 & \textbf{570.89} & \textbf{593.42} & 621.27 & 677.03 \\ \hline 
c75 & \textbf{962.79} & 1027.08 & 1003.20 & 1013.47 & 981.57 & \textbf{1013.45} & 998.72 & 1066.00 \\ \hline 
c100 & 1000.98 & 1013.03 & 987.65 & 1012.30 & \textbf{961.10} & \textbf{987.59} & 979.95 & 1066.44 \\ \hline 
c100b & 899.05 & 931.35 & 932.35 & 943.05 & 881.92 & 900.94 & \textbf{823.23} & \textbf{831.72} \\ \hline 
c120 & 1328.54 & 1418.13 & 1272.65 & 1451.60 & 1303.59 & 1390.58 & \textbf{1068.46} & \textbf{1100.28} \\ \hline 
c150 & 1412.03 & 1461.55 & 1370.33 & 1394.77 & 1348.88 & 1386.93 & \textbf{1223.15} & \textbf{1323.35} \\ \hline 
c199 & 1778.56 & 1843.06 & 1717.31 & 1757.02 & 1654.51 & 1758.51 & \textbf{1533.68} & \textbf{1601.22} \\ \hline 
\hline 
f71 & 304.51 & 323.91 & 311.33 & 320.00 & 301.79 & \textbf{309.94} & \textbf{288.72} & 323.85 \\ \hline 
f134 & 16063.65 & 16671.17 & 15557.82 & 16030.53 & 15528.81 & 15986.84 & \textbf{12134.30} & \textbf{12473.32} \\ \hline 
\hline 
tai75a & 1822.38 & 1871.46 & 1832.84 & 1880.87 & \textbf{1782.91} & \textbf{1856.66} & 1899.17 & 1980.37 \\ \hline 
tai75b & \textbf{1433.98} & 1533.63 & 1456.97 & \textbf{1477.15} & 1464.56 & 1527.77 & 1515.71 & 1582.37 \\ \hline 
tai75c & 1505.06 & 1558.70 & 1612.10 & 1692.00 & \textbf{1440.54} & \textbf{1501.91} & 1526.75 & 1657.88 \\ \hline 
tai75d & 1434.18 & 1458.93 & 1470.52 & 1491.84 & \textbf{1399.83} & \textbf{1422.27} & 1426.39 & 1456.08 \\ \hline 
tai100a & \textbf{2223.04} & \textbf{2290.05} & 2257.05 & 2331.28 & 2232.71 & 2295.61 & 2344.78 & 2532.85 \\ \hline 
tai100b & 2221.58 & 2263.46 & 2203.63 & 2317.30 & \textbf{2147.70} & \textbf{2215.39} & 2221.67 & 2344.13 \\ \hline 
tai100c & \textbf{1518.08} & \textbf{1541.25} & 1660.48 & 1717.61 & 1541.28 & 1622.66 & 1580.00 & 1676.72 \\ \hline 
tai100d & 1870.50 & 2004.78 & 1952.15 & 2087.96 & \textbf{1834.60} & \textbf{1912.43} & 1888.07 & 2000.07 \\ \hline 
tai150a & 3508.09 & 3570.51 & 3436.40 & 3595.40 & \textbf{3328.85} & \textbf{3501.83} & 3607.78 & 3763.44 \\ \hline 
tai150b & 3019.90 & 3120.57 & 3060.02 & \textbf{3095.61} & \textbf{2933.40} & 3115.39 & 3070.44 & 3226.49 \\ \hline 
tai150c & 2959.58 & 3065.73 & 2735.39 & 2840.69 & \textbf{2612.68} & 2743.55 & 2614.59 & \textbf{2725.45} \\ \hline 
tai150d & 3008.30 & 3175.37 & 3138.70 & 3233.39 & \textbf{2950.61} & \textbf{3045.16} & 3081.26 & 3200.29 \\ \hline 
tai385 & 40238.00 & 41319.39 & 33062.06 & 35188.99 & - & - & \textbf{31876.05} & \textbf{33786.47} \\ \hline 
\hline 
sum & 91115.53 & 94081.97 & 83632.73 & 87495.92 & 49202.73 & 51088.83 & \textbf{78324.14} & \textbf{82395.82} \\ \hline 

 \end{tabular}
} \end{center}
\end{table*}

The comparison of~the~best performing version of~each approach is presented
in Table~\ref{tab:comparable}, while performance of~each of~the~methods for various parameters
setting is presented in~Fig.~\ref{fig:algorithm.performance}. Tree and MCTree algorithms have been run with the~working day divided into: $40$, $50$, $100$ and $200$ time slices (see Fig.~\ref{fig:tree} and Fig.~\ref{fig:mctree}).
{Dividing working day into $200$ time slices results in~running optimization nearly every
time when the~new request arrive even in~the~largest benchmarks. Additionally, the~total
processing time for the~largest benchmarks is close to~the~limit defined by the~literature approaches.}
MCTree+PSO and 2MPSO algorithm have been both run with the~working day divided
into $40$ time slices (the number tuned for the~2MPSO algorithm). PSO in~the~MCTree+PSO has been run with
the following pairs of~swarm size and iterations limit: $1 \times 1$, $2 \times 1$, $2 \times 14$,
$3 \times 21$, $6 \times 42$ and $7 \times 49$ (see Fig.~\ref{fig:mctreepso}). PSO in~the~2MPSO has been run 
with
the following pairs of~swarm size and iterations limit: $1 \times 1$, $2 \times 14$,
$3 \times 21$ and $4 \times 28$ (see Fig.~\ref{fig:mpso}).
{The largest swarm sizes and number of~iterations have been chosen
in order to~stay within the~average total processing time of~the~Tree with $200$ time slices
(see Fig.~\ref{fig.time.2mpso.vs.mctree}).}
{The smaller swarm sizes and number of~iterations were tested in
order to~find algorithms and configurations for the~problems were the~time limit
for processing becomes a~crucial constraint (possibly due to~the~large
number of~requests).}

The average results and the~average computational effort for each of~the~tested parameters
set is depicted on the~log--log plot of~time to~result trade-off in~Fig.~\ref{fig.time.2mpso.vs.mctree}.
A few chosen cases were additionally depicted in~Fig.~\ref{fig:time.compare}
presenting computational effort for the~benchmarks with different number of~requests.
The regression curves fitted for the~time of~computations as a~linear function of~$m^2log(m)$,
explained at~least 95\% of~the~variance for each of~the~algorithms, thus confirming
the theoretical time complexity of~all the~approaches being $o(m^2log(m)$) due to~
the computations needed for the~clustering (sorting the~edges in~discrete (MC)Tree case).

From both Fig.~\ref{fig.time.2mpso.vs.mctree} and Tab.~\ref{tab:comparable}
we could observe, that the~MCTree+PSO performed much better than plain MCTree
(which performed better than plain Tree algorithm).
Therefore, MCTree+PSO is compared with literature results in~Table~\ref{tab:external}.
The size of~the~swarm (7) and the~number of~iterations (49) for the~PSO part
of the~method were chosen in~such way, that the~algorithm would not exceed
75 seconds for \texttt{c199} benchmark and 500 seconds for \texttt{tai385}.
The reason being our Intel Core i7@3.4GHz machine is about 10 times faster than the~Pentium IV@2.8GHz
used for the~GA2007 computations (with 750 seconds limit) and about 3 times
faster then the~Intel Core i5@2.4GHz used for GA2014 and ACOLNS (which were computed with 1500 seconds limit).
While  Table~\ref{tab:external} presents the~literature results achieved
only by the~methods using time as the~optimization process limit,
the best known values for the~benchmarks can be found at~the~research project~\cite{DVRP:Project}
website\footnote{\url{http://www.mini.pw.edu.pl/~mandziuk/dynamic/?page_id=67#results}}.

\section{Discussion and Conclusions}
\label{sec:discussion}
{
Proposed MCTree algorithm improved the~solution which could be achieved
by the~same heuristic algorithm Tree without the~generation of~the~artificial requests.
Further improvement of~the~average results was possible with the~MCTree+PSO hybrid.
}
MCTree+PSO approach for solving DVRP proved to~be beneficial over 2MPSO algorithm
for small time budget ({less then 10 seconds on average for this particular
benchmark set on the~Intel Core i7 machine}).
Additionally, it was competitive against both the 2MPSO and GA2007
algorithms, within the~larger processing time limit defined by the~GA2007.
It achieved to~get better average results against 2MPSO
for the~$10$ (out of~$22$) benchmark instances and against GA2007 for the~$6$ (out of~$21$).
The average routes computed by MCTree were 1.0144 times longer than GA2007,
and 1.0126 times shorter than GA2014 and 1.0178 shorter than ACOLNS
while maintaining strict time limit on the~largest benchmarks.
Probably, this results could be improved for smaller benchmarks (and thus on the~average)
if the~MCTree+PSO algorithm used the~time as the~limit for the~processing
for the~2MPSO.

It is important to~note that the~MCTree+PSO with just single evaluation
by the~continuous approach has the~best average processing time
and should be able to~process DVRPs with around $10 000$ requests
in $8$ hours on a~single PC (conclusion drawn from the~regression curves depicted in~Fig.~\ref{fig:time.compare}). The~largest (3 particles
and 21 iterations) MCTree+PSO performing better then plain 2MPSO should be able to~process problems with around $5 400$ requests in~that time
and 2MPSO with $2$~particles and $14$~iterations (having similar quality of~results)
problems with around $4 700$ requests.

It could be also observed that even such simple generation of~the~artificial requests
improved the~performance of~the~clustering based on the~Kruskal algorithm
and route construction based on the~2--OPT algorithm.

The proposed approach (generating of~artificial requests) might be
improved by using another clustering algorithm to~solve the~assignment
of the~requests. Another area for further development is more sophisticated
generation of~artificial requests for a~closer resemblance of~their
spatial distribution.

\section*{Acknowledgments}
The research was financed by the~National Science Centre
in Poland grant number DEC-2012/07/B/ST6/01527~\cite{DVRP:Project} and by the~research 
fellowship within "Information
technologies: Research and their interdisciplinary applications"
agreement number POKL.04.01.01-00-051/10-00.

\bibliographystyle{splncs}
\bibliography{References}

\end{document}